\documentclass[sigconf]{acmart}
\usepackage{graphicx}
\usepackage{subcaption}
\usepackage{bbding} 
\usepackage[linesnumbered,ruled,vlined]{algorithm2e}
\usepackage{amsmath}
\usepackage{booktabs}
\usepackage{multirow}
\usepackage{array}

\ifdefined \GramaCheck
  \newcommand{\CheckRmv}[1]{}
  \newcommand{\figref}[1]{Figure 1}%
  \newcommand{\tabref}[1]{Table 1}%
  \newcommand{\secref}[1]{Section 1}
  \renewcommand{\equref}[1]{Equation 1}
\else
  \newcommand{\CheckRmv}[1]{#1}
  \newcommand{\figref}[1]{Figure ~\ref{#1}}%
  \newcommand{\subfigref}[2]{Figure \ref{#1}(\subref{#2})}

  \newcommand{\tabref}[1]{Table~\ref{#1}}%
  \newcommand{\secref}[1]{Section~\ref{#1}}
  \renewcommand{\eqref}[1]{Equation~(\ref{#1})}
\fi

\def\eg{\emph{e.g.,~}}

\newcommand{\ourM}{{LightCL}}
\newcommand{\sArt}{{state-of-the-art~}}
\newcommand{\CL}{{CL}}

\AtBeginDocument{%
  }

\copyrightyear{2025}
\acmYear{2025}
\setcopyright{acmlicensed}\acmConference[ASPDAC '25]{30th Asia and South Pacific Design Automation Conference}{January 20--23, 2025}{Tokyo, Japan}
\acmBooktitle{30th Asia and South Pacific Design Automation Conference (ASPDAC '25), January 20--23, 2025, Tokyo, Japan}
\acmDOI{10.1145/3658617.3697784}
\acmISBN{979-8-4007-0635-6/25/01}

\begin{document}
\newcommand{\equalcontrib}{\textsuperscript{*}}
\newcommand{\correspondingauthor}{\textsuperscript{$\dagger$}}

\title{The Name of the Title Is Hope}

\title{\ourM: Compact Continual Learning with Low Memory Footprint For Edge Device}

\author{Zeqing Wang\equalcontrib}
\affiliation{%
    \institution{Xidian University}
    \city{Xi’an}
    \country{China}}
\email{zeqing.wang@stu.xidian.edu.cn}

\author{Fei Cheng\equalcontrib\correspondingauthor}
\affiliation{%
    \institution{Xidian University}
    \city{Xi’an}
    \country{China}}
\email{chengfei@xidian.edu.cn}

\author{Kangye Ji}
\affiliation{%
    \institution{Xidian University}
    \city{Xi’an}
    \country{China}}
\email{kangye.ji@stu.xidian.edu.cn}

\author{Bohu Huang}
\affiliation{%
    \institution{Xidian University}
    \city{Xi’an}
    \country{China}}
\email{huangbo@mail.xidian.edu.cn}

% \author{Zeqing Wang\equalcontrib, Fei Cheng\equalcontrib\correspondingauthor, Kangye Ji, Bohu Huang}
% \affiliation{
% \institution{Xidian University}
% \city{Xi'an}
% \country{China}
% }

% \email{
% {zeqing.wang, kangye.ji}@stu.xidian.edu.cn; 
% chengfei@xidian.edu.cn; 
% huangbo@mail.xidian.edu.cn
% }

\renewcommand{\shortauthors}{Zeqing Wang, Fei Cheng, Kangye Ji, and Bohu Huang} % !! %%%
\renewcommand\arraystretch{0.4}

\begin{abstract}
Continual learning~(CL) is a technique that enables neural networks to constantly adapt to their dynamic surroundings. Despite being overlooked for a long time, this technology can considerably address the customized needs of users in edge devices.
Actually, most \CL~ methods require huge resource consumption by the training behavior to acquire generalizability among all tasks for delaying forgetting regardless of edge scenarios. Therefore, this paper proposes a compact algorithm called \ourM, which evaluates and compresses the redundancy of already generalized components in structures of the neural network.
Specifically, we consider two factors of generalizability, learning plasticity and memory stability, and design metrics of both to quantitatively assess generalizability of neural networks during \CL. This evaluation shows that generalizability of different layers in a neural network exhibits a significant variation.
Thus, we \textit{Maintain Generalizability} by freezing generalized parts without the resource-intensive training process and \textit{Memorize Feature Patterns} by stabilizing feature extracting of previous tasks to enhance generalizability for less-generalized parts with a little extra memory, which is far less than the reduction by freezing.
Experiments illustrate that \ourM~ outperforms other state-of-the-art methods and reduces at most $6.16$$\times$ memory footprint. We also verify the effectiveness of \ourM~ on the edge device.

\end{abstract}



% \keywords{Do, Not, Us, This, Code, Put, the, Correct, Terms, for,
  

\maketitle

\renewcommand{\thefootnote}{\fnsymbol{footnote}}
\footnotetext[1]{Equal contribution}
\footnotetext[2]{Correspondence to chengfei@xidian.edu.cn}
\footnotetext{Our code is available at: https://github.com/INV-WZQ/LightCL}

\section{Introduction}\label{sec:intro}
The wave of deep learning has significantly boosted the deployment of neural networks from cloud platforms to edge devices \cite{TinyML-survey,256kb}. As the dynamic realistic environment requires the adaptive ability of models, a potential technique known as continual learning (CL) empowers these models to continually acquire, update, and accumulate knowledge on the device from the data by the training behavior. Although powerful cloud platforms can leverage \CL~ to adapt to service expansion, \eg personalized recommendation systems \cite{cloud}, the similar and intense personalized requirements for edge devices are almost disregarded by academia and industry. This overlook stems from scarce resources in edge devices, such as limited computation, constrained storage, and insufficient memory, which significantly hinders the further application of \CL. Hence, the efficient \CL~on the edge device becomes a considerable challenge. 
	
Most existing \CL~ methods ignore edge scenarios and only concentrate on addressing the catastrophic forgetting problem \cite{forgetting, EWC} -- model training on a new task tends to "forget" the knowledge learned from previous tasks. Specifically, these methods explore two factors of \textit{generalizability} \cite{2023survey}, learning plasticity and memory stability, and make a trade-off between them to delay forgetting. Note that learning plasticity denotes the adaptating ability to the new task, while memory stability denotes the memory capacity of previous tasks. Unfortunately, most of \CL~ methods \cite{DER, er, FDR, EWC, PNN} simultaneously train on the trace of historical information and new data for generalizability.
It is well known that the training overhead of the neural network is already significant. Meanwhile, learning from history in \CL~ requires additional resources for the training process. Accordingly, the casual \CL~methods are unsuitable for resource-intensive scenarios.
To mitigate this problem, compression techniques are inevitably introduced to the on-device \CL. Especially, sparse learning \cite{prune-retrain} has become a popular paradigm of the efficient \CL~ \cite{SparCL, EsaCL}. 
Although they \cite{SparCL, EsaCL} greatly reduce computation by removing unimportant weights, intermediate feature maps cannot be effectively reduced, leading to an enormous consumption of memory footprint during training. 

Before introducing our work, we will give a comprehensive analysis of the challenges encountered.
\textit{Firstly}, the forgetting problem is the first and foremost issue of \CL, whether on edge devices or cloud platforms. Most CL methods train on historical information and new data together for generalizability to delay forgetting. This training behavior requires more significant resources than usual, including more intermediate feature maps and gradients in memory, as well as additional computations for gradient calculations. This greatly hinders the efficiency. 
\textit{Secondly}, the efficiency of CL heavily influences its prevalence in edge devices. 
The computation and memory footprint are two main factors that hinder the deployment. 
Meanwhile, memory rather than computation has become the primary bottleneck in AI applications, and the time taken for memory transfers greatly limits computational capacity \cite{memorywall}, which impedes the process of overcoming forgetting during \CL.

\textit{Actually}, the above two challenges are \textit{not} orthogonal. The former inevitably interferes with the latter, and vice versa. This tangle prompts us to solve two challenges together. 
Considering the neural network has the potential to train in new tasks with the help of previous knowledge during \CL, it absolutely has the redundancy of generalizability. 
In other words, \textit{if we could evaluate the distribution of generalizability in the neural network during \CL, we can compress the training redundancy with the over-strong generalizability and use just a few extra resources to maintain memory in less-generalized parts, which hopefully gives us a more favorable and efficient CL method on edge devices.}

\begin{figure}[ht]
    \centering
    \includegraphics[scale=0.05]{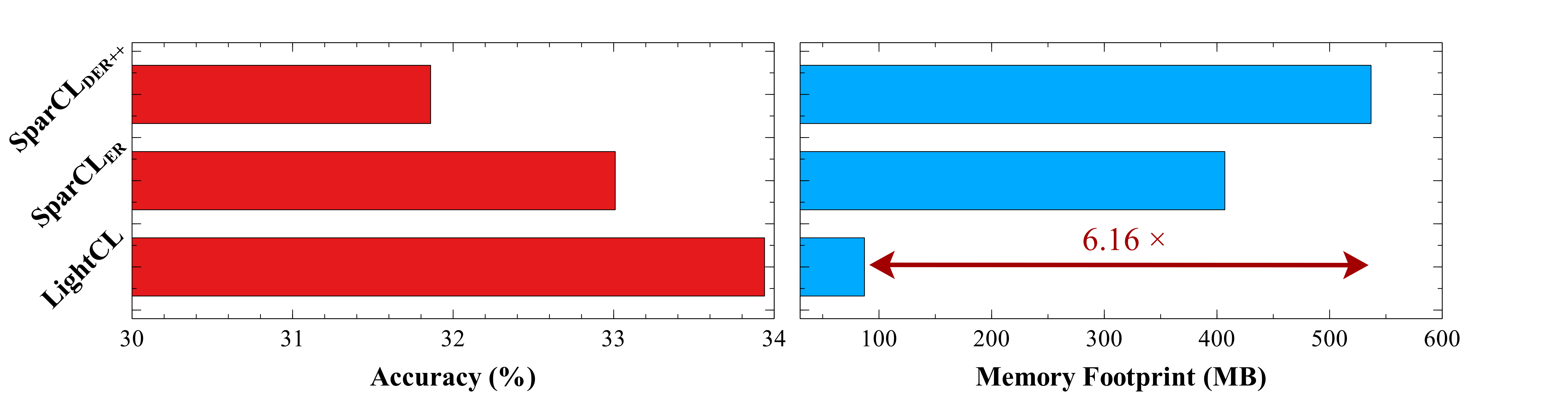}
    \caption{Evaluation of accuracy and memory footprint on Split Tiny-ImageNet \cite{Imagenet} under the TIL setting with sparsity ratio $90$\%. The performance of LightCL outperforms the \sArt efficient CL method SparCL \cite{SparCL} and reduces at most $6.16$$\times$ memory footprint compared with SparCL$_{DER++}$.}
    \label{fig:clear_comparison}
\end{figure}
Therefore, we start with the two factors of generalizability, learning plasticity and memory stability, and design their metrics to quantitatively assess the generalizability of neural networks during CL. This evaluation demonstrates that
the generalizability of different layers in a neural network exhibits a significant variation. 
In particular, lower and middle layers have more generalizability, and deeper layers are the opposite. According to this conclusion, we \textsl{Maintain Generalizability} by easily freezing lower layers to delay forgetting problem and greatly minimize the consumption of memory footprint and computation. 
Furthermore, as for less-generalized parts in deeper layers, we regulate their feature maps, necessitating fewer resources than weights, to \textsl{Memorize Feature Patterns} of previous tasks for the promotion of generalizability.  
The resource consumption of \textsl{Memorize Feature Patterns} is significantly less than its reduction of \textsl{Maintain Generalizability}. Thus, our method can achieve great performance with limited resources. Since our method achieves high efficiency only by exploring the network structure in CL, it can be easily integrated with other compression techniques. As shown in \figref{fig:clear_comparison}, with the same sparsity ratio, the performance of LightCL outperforms the \sArt efficient CL method SparCL and reduces at most $6.16$$\times$ memory footprint.

In summary, our work makes the following contributions:
\begin{itemize}
    \item It is the \textit{first} study to propose two new metrics of learning plasticity and memory stability to quantitatively evaluate generalizability for efficient \CL. We conclude that lower and middle layers have more generalizability, while deeper layers hold the opposite.
    
    \item Based on the evaluation, \ourM~ efficiently overcomes catastrophic forgetting through \textsl{Maintain Generalizability} and \textsl{Memorize Feature Patterns.} 
    
    \item Experiments illustrate that \ourM~ outperforms \sArt \CL~ methods in delaying the forgetting problem, and the memory footprint of \ourM~ is reduced at most $6.16$$\times$. 
\end{itemize}

The rest of the paper is organized as follows. Related work is reviewed in \secref{sec:relate}. In \secref{sec:CL-setting}, we introduce the \CL~ setting related to this paper. We design the metrics and evaluate the generalizability in \secref{sec:analysis}. According to the evaluation, we propose \ourM~ in \secref{sec:Proposed Architecture}. Experimental analysis and conclusion are illustrated in \secref{sec:experiment} and \secref{sec:conclusion}, respectively.
\section{RELATED WORK}\label{sec:relate}
	\noindent \textbf{Continual Learning Methods.  }
 \CL~ methods can be categorized into replay methods \cite{DER, er, FDR, remind, Co2l}, regularization-based methods \cite{EWC, LWF, SI, MAS}, and architecture-based methods \cite{Packnet, PNN, LPS}. Replay methods address catastrophic forgetting by replaying a subset of old data stored in storage. They must continually store samples when they meet new tasks, and bring extra computation and memory footprint for training on old data. 
 Regularization-based methods are designed to restrict the updates of neural networks through regularization, preserving valuable information about previous tasks. Most of them \cite{EWC, SI, MAS} limit changes of parameters that are important to old tasks. This manner leads to extra consumption for recording the importance of every parameter. 
 Architecture-based methods can mitigate forgetting by distributing different model structures to each task. Similar work \cite{PNN} instantiates a new network for each new task, while others \cite{LPS} divide a network into task-specific parts through pruning. Hence, most of them grow the model size continually or train on a whole dense model to fit new tasks, which causes significant resource overhead during the \CL~ process. Meanwhile, they must acquire the identity of the task before an evaluation, which is unsuitable for edge scenarios.

\noindent \textbf{Efficient Learning. }Many efficient algorithms aim at the deployment of neural networks on edge devices, such as pruning \cite{deepcompression, filterspruning}, quantization \cite{Quantization}, etc. Pruning compresses networks by discarding redundancies of weights, while quantization reduces the bit-width of weights. These methods mainly accelerate the inference procedure on edge \cite{inference}, and degrade the ability of compressed models to accommodate dynamic environments. This is because a large training burden hinders deployment development, such as the extensive use of intermediate feature maps and gradients, which increases memory footprint and requires more training FLOPs for computation. Although \cite{256kb} can train in resource-limited settings, they focus more on training a single task and do not consider catastrophic forgetting between multiple tasks. To cope with catastrophic forgetting of \CL~ in the edge device scenario, SparCL \cite{SparCL}, combined with the dynamic pruning method, greatly reduces training FLOPs and delays the forgetting problem. However, SparCL still occupies massive memory footprints for intermediate feature maps. 
	
\section{CL Setting}\label{sec:CL-setting}
	In the \CL~ setting, tasks $\{1, 2,..., T\}$ are executed sequentially during training, and data from previous tasks are inaccessible when training on the new task. Specifically, the \textsl{t}-th task represents $\mathcal D_t = \{(x_t^i, y_t^i)\}^{n_t}_{i=1}$, where $n_t$ is the number of samples of \textsl{t}-th task and $(x_t^i, y_t^i)$ is input-label pair. $\theta_{i}^t$ represents the \textsl{i}-th layer of the neural network after (or during) training \textsl{t}-th task. The goal of \CL~ is to let the AI model perform well on all tasks when learning a new task. Overall performance is typically evaluated by average accuracy (AA) \cite{2023survey, SparCL} on all tests after training. AA is defined as $AA = \frac{1}{T} \sum_{i=1}^T a_i(\theta^T)$, where $a_i(\theta)$ represents the testing accuracy of model $\theta$ on \textsl{i}-th task. There are two main \CL~ scenarios \cite{SparCL}: 1) Task incremental learning (TIL), the identity $t$ of the task $\mathcal D_t$ is available during training and testing; 2) Class incremental learning (CIL), the task identity $t$ are only available during training. This paper evaluates \CL~ methods under both TIL and CIL scenarios.
\section{Analysis of Generalizability}\label{sec:analysis}
In this section, we analyze the distribution of generalizability for the neural network during \CL. Training consumption on generalized components in the neural network can be a kind of redundancy since certain generalized components contain previous knowledge while already have the ability to fit new tasks prior to encountering them. This is the motivation for evaluating generalizability. 
As \cite{2023survey} points out that generalizability can be utilized for both previous and new tasks during \CL, memory stability (MS) and learning plasticity (LP) are two different characteristics of it. MS denotes the loss of previous knowledge, and LP denotes the adaptation to new knowledge. Therefore, we need to analyze the generalizability from these two perspectives.
Although similar work \cite{anatomy} makes a deep analysis in continual learning, it focuses more on analyzing forgetting parts in the neural network. In fact, forgetting means strong learning plasticity and weak memory stability, so generalizability and forgetting are not orthogonal and we cannot infer generalizability from the previous work \cite{anatomy}. Hence, we need to design two metrics of LP and MS to quantitatively evaluate generalizability.

Intuitively, to analyze MS and LP in specific structures of neural networks, the possible way is to analyze the effectiveness of specific structures on previous and new tasks. Through testing on previous tasks, we are able to evaluate the loss of previous knowledge in certain structures. By testing on new tasks, we can also evaluate their adapting ability to new knowledge.
Specifically, we quantitatively evaluate the above metrics of each layer of the neural network. We replace the parameters of \textsl{i}-th layer in the model $\theta^T$ with the parameters of the same layer from the previous model $\theta^{T-1}$. The new model $\theta^{new_i}$ can be described as
\begin{equation}
    \begin{gathered}[t]
        \theta^{new_i} = 
        \begin{cases}
            \theta^{new_i}_j = \theta^T_j & \text{if } j \neq i, \\
            \theta^{new_i}_i = \theta^{T-1}_i & \text{otherwise.}
        \end{cases}
    \end{gathered}
    \label{equ:analysis_1}
\end{equation}
Then, we separately test the accuracy of $\theta^{new_i}$ on \textsl{(T-1)}-th task and \textsl{T}-th task. Two metrics of LP and MS can be described as 
\begin{equation}
    \begin{gathered}[t]
        \begin{cases}
            MS = a_{T-1}(\theta^{new_i}) - a_{T-1}(\theta^{T}),     \\
            LP = a_{T}(\theta^{new_i}) - a_{T}(\theta^{T}).
        \end{cases}
    \end{gathered}
    \label{equ:analysis_2}
\end{equation}
%MS measures the loss of previous knowledge and LP measures adaptation to new knowledge. 
If the value of MS is high, it indicates that the previous knowledge in \textsl{i}-th layer has been lost to a large extent, suggesting that the memory stability of \textsl{i}-th layer is weak. The low value of LP indicates that the previous \textsl{i}-th layer does not have a good adaptation for the new tasks, meaning that the previous \textsl{i}-th layer has weak plasticity. 

\begin{figure}[htbp]
    \centering
    \begin{subfigure}{0.227\textwidth}
        \centering
        \includegraphics[width=1.\linewidth]{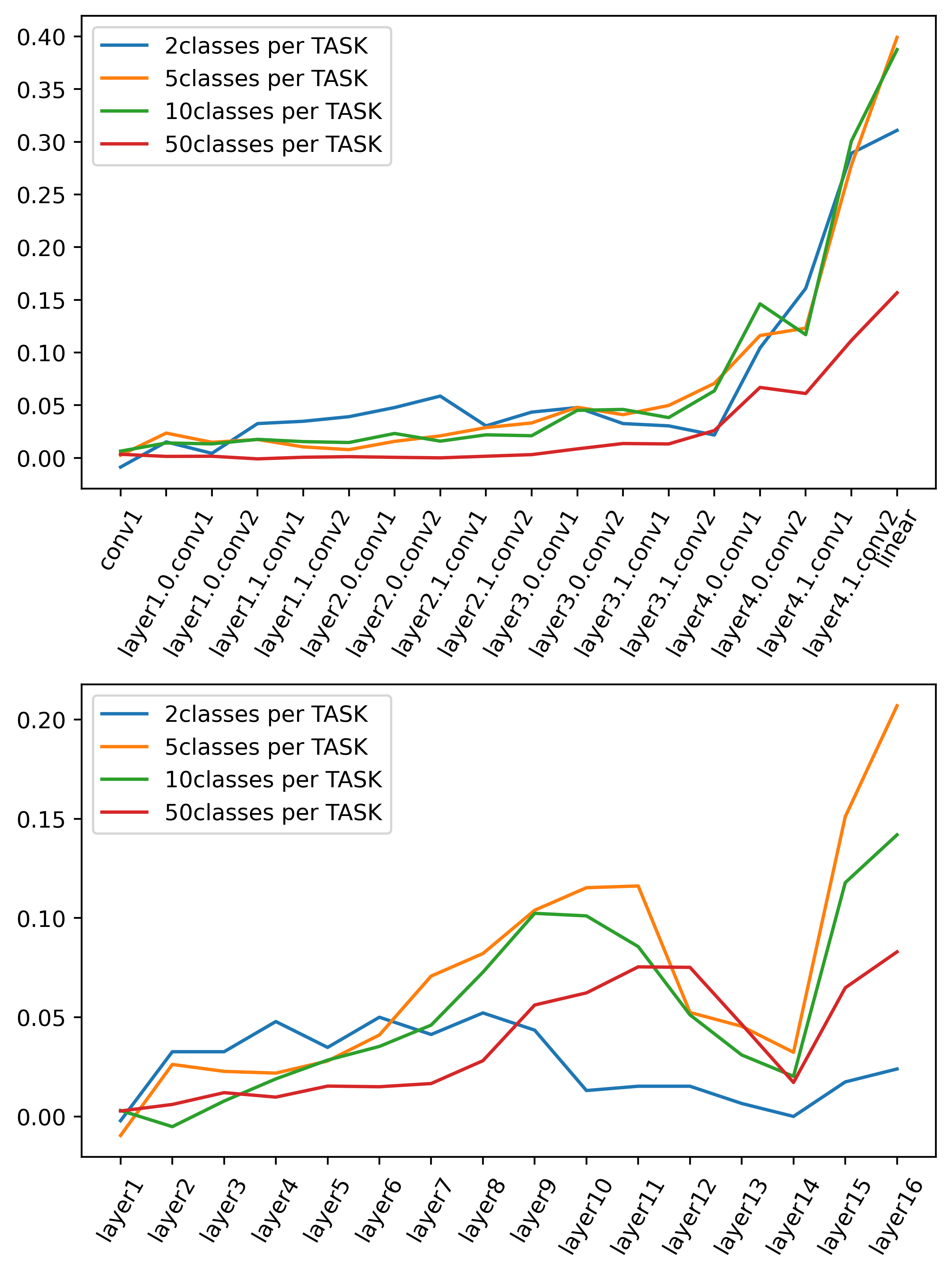}
        \caption{Evaluation of MS\label{fig:analysis_1}}
    \end{subfigure}
    \quad
    \begin{subfigure}{0.227\textwidth}
        \centering
        \includegraphics[width=1.\linewidth]{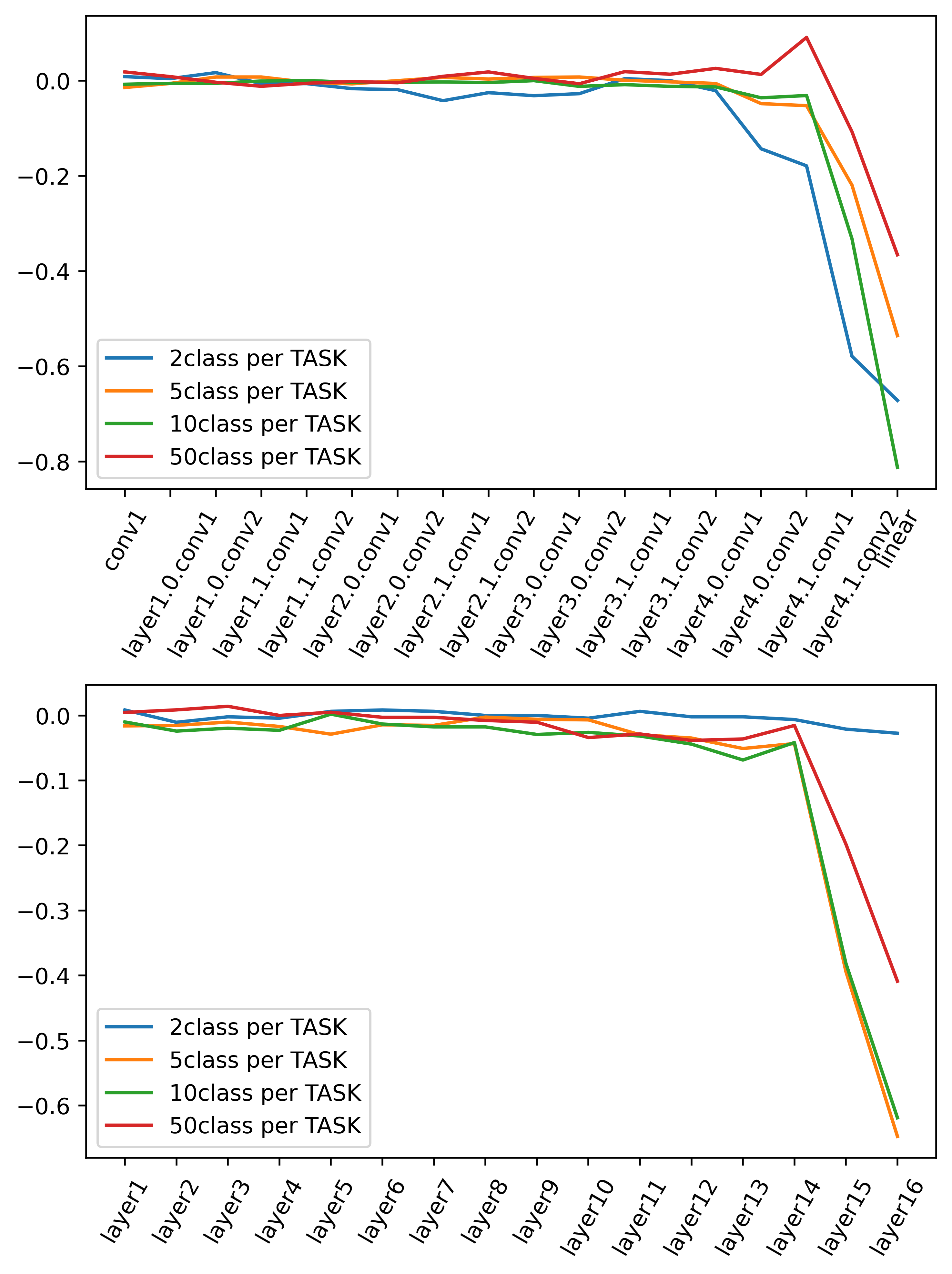}
        \caption{Evaluation of LP\label{fig:analysis_2}}
    \end{subfigure}
    \caption{Evaluation of MS and LP on Split-TinyImageNet dataset with ResNet-18 (first row) and VGG16 (second row) under CIL setting. The X-axis is the index of each layer in the certain network. We operate two consecutive tasks for \CL, containing the same number of classes, to evaluate the metrics of each layer. Different colors of lines represent different tests with different class numbers. All tests run $5$ times.}
    \label{fig:analysis}
\end{figure}

In order to make our experiment more reasonable, we evaluate the above metrics using different class numbers for multiple tasks across $5$ runs with two typical backbones, ResNet-18 and VGG16 \cite{VGG}. As shown in \figref{fig:analysis}, different numbers of classes exhibit a similar phenomenon that, in lower layers, the values of MS are small while the values of LP are high. In other words, the lower layers have strong memory stability and learning plasticity, which exhibit strong generalizability. The middle layers in ResNet-18 display strong generalization ability, while the counterpart of VGG16 exhibits poor memory stability, shown in the second row of \subfigref{fig:analysis}{fig:analysis_1}. As shown in \subfigref{fig:analysis}{fig:analysis_2}, high value in LP of middle layers is important to note that middle layers already exhibit learning plasticity before being trained on the new task, indicating that learning the new task will impair the stability of the middle layers and degrade the existing generalizability. This phenomenon also proves the opinion that forgetting parts is not orthogonal to generalized parts since they may display different features during learning process like middle layers. For the deeper layers, the values of MS are higher and the values of LP are lower. This
suggests that the deeper layers are less-generalizability and forgetting parts. From the above analysis, we conclude that \textbf{during \CL, lower and middle layers have strong generalizability, and deeper layers have less generalizability.} This conclusion guides the design of \ourM~ in \secref{sec:Proposed Architecture}.

\section{Proposed Architecture}\label{sec:Proposed Architecture}

\begin{figure}[ht]
    \centering
    \includegraphics[scale=0.31]{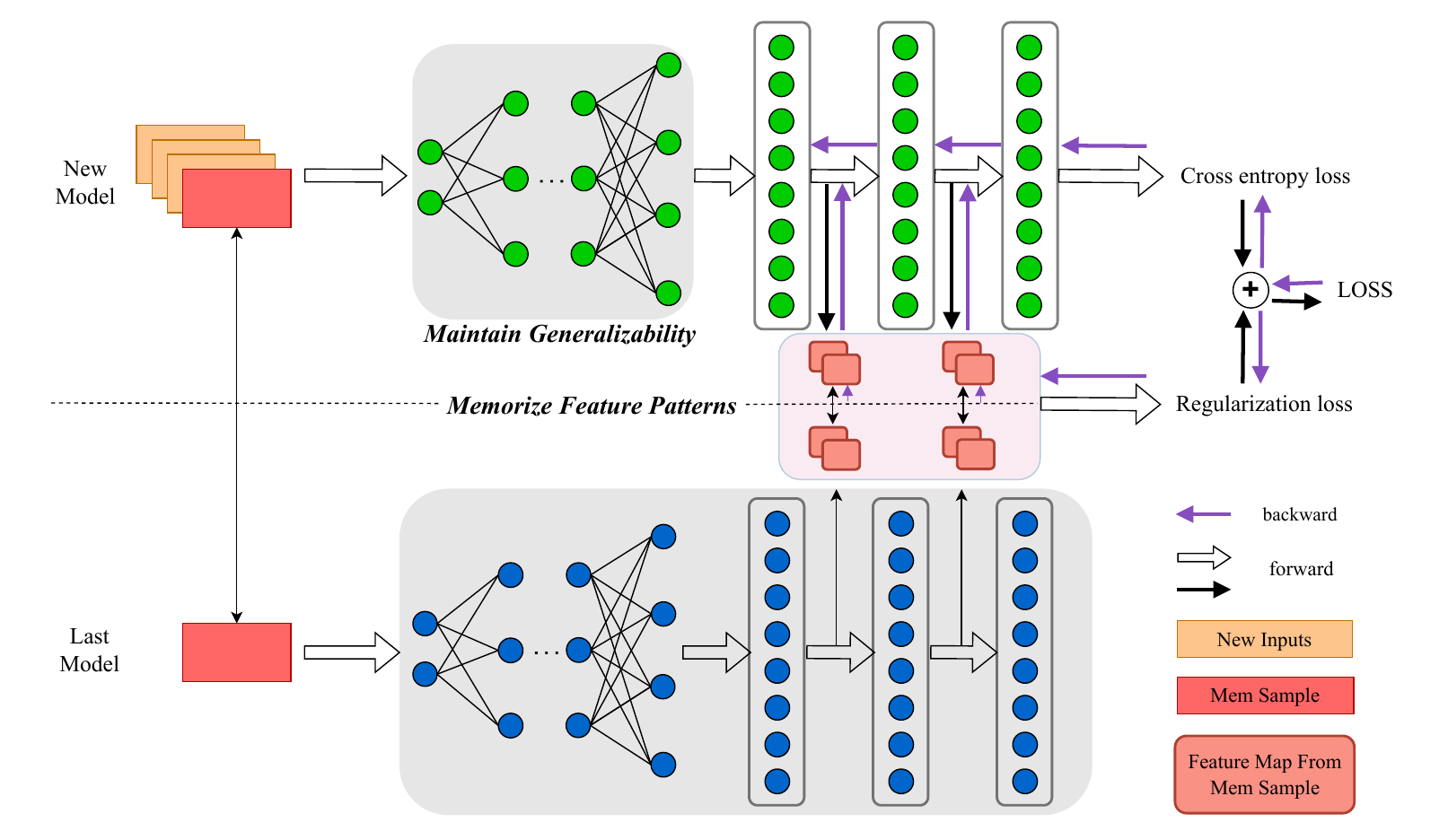}
    \caption{Illustration of each component in \ourM. Before training a new task, we obtain feature map standards as references for later regulation and store them in memory, as shown in the bottom part of the figure. During the training process, shown in the top part of the figure, we \textsl{Maintain Generalizability} by freezing lower and middle layers, and \textsl{Memorize Feature Patterns} by regulation between current feature maps and stored feature map standards.}
    \label{fig:pipeline}
\end{figure}

 \ourM, as shown in \figref{fig:pipeline}, efficiently delays forgetting through \textsl{Maintain Generalizability} and \textsl{Memorize Feature Patterns.} All these methods are based on the discovery of generalizability in \secref{sec:analysis}. We first introduce \textsl{Maintain Generalizability} in \secref{sec:Maintaining Generalizability} and then illustrate \textsl{Memorize Feature Patterns} in \secref{sec:Memorize Feature Patterns}.

\subsection{Maintain Generalizability}\label{sec:Maintaining Generalizability} 
    According to the discovery in \secref{sec:analysis} that lower and middle layers have strong generalizability, the training consumption on them can be a kind of redundancy. Hence, we easily freeze them to \textsl{maintain generalizability} without the training process, illustrated in the gray background section at the top of the \figref{fig:pipeline}. Specifically, we must pretrain a network, freeze its lower and middle layers, and finally deploy it on the edge device for \CL. Training the \textsl{t}-th task can be formulated as
	\begin{equation}
		\label{equ:maintain_1}
		\begin{array}{cl}
			\theta_{i}^t \leftarrow \theta_{i}^t + \alpha \nabla_{\theta_{i}^t} \mathcal L_c(f(\theta^t,x_t),y_t) \quad
			\text{s.t.} \quad i > \textrm{FZ},
		\end{array}
	\end{equation}
    where FZ is the last layer index of frozen layers, $\mathcal L_c$ is the cross-entropy loss function, and function $f$ represents forward pass. For example, we set FZ as layer3.1.conv2 based on the discovery when we utilize ResNet-18 as the backbone network. This setting will be kept for experiments in \secref{sec:experiment}.
    
 \textsl{Maintain Generalizability} greatly reduces resource consumption. Firstly, as frozen layers do not participate in backpropagation, there is no need to compute and store gradients and feature maps during training, which greatly reduces memory usage. Secondly, the complexity of the computation graph shrinks, so backpropagation requires less computation. This alleviates the burden on hardware, thereby improving computational efficiency and shortening training time. Lastly, frozen layers decrease data transfer in memory during backward propagation, hence reducing latency and improving data flow efficiency. %Meanwhile, frozen parts are able to keep their knowledge completely, and the forgetting problem can also be delayed. 
 Notably, this high-efficiency method is able to delay the forgetting procedure without diminishing the knowledge of counterparts. As evidenced by the experiments in \secref{sec:compare_pretrain}, with the help of \textsl{Maintain Generalizability}, \ourM~ reduces $6.16$$\times$ memory footprint and is competitive beyond the efficient \CL~ method SparCL in FLOPs. Meanwhile, it has greatly delayed forgetting by the seamless integration with most \CL~ methods in \secref{sec:Ablation_maintain}, which illustrates the powerful scalability of our method. 

\subsection{Memorize Feature Patterns}\label{sec:Memorize Feature Patterns}
	Based on the conclusion (\secref{sec:analysis}) that deeper layers are less generalized and forgetting parts, we propose an efficient regulation method to memorize previous knowledge in deeper layers, shown in the pink background in the middle of the \figref{fig:pipeline}. We go through the following three subsections to illustrate it.
    
    \subsubsection{How to Memorize Efficiently}\label{sec:How to Regulate}
	Forgetting occurs because the gradient update of the current task does not consider generalizability among all tasks. Restricting the updating of vital weights of previous tasks, as the regularization-based method \cite{EWC} does, is the direct way to achieve our goal. \cite{256kb} points out that the deeper the layers, the more the memory footprint occupied by weights, while the size of feature maps becomes smaller and smaller. Our idea is that since both feature maps and weights can reflect the output patterns of the model, we can memorize previous feature-extracting patterns by regulating important feature maps to overcome forgetting with a little extra consumption.	
	\subsubsection{How to Select Reference}
    The intuitive idea is to regulate changes in vital feature maps produced by each sample of previous tasks during training for a new task. 
    However, it is impossible to store samples continuously in resource-limited scenarios. As the number of tasks increases, so does the number of samples required. Alternatively, we regulate feature maps from the current sample instead of previous samples because regulating output patterns of feature maps is unrelated to the information of samples. The only effort is to recognize feature maps that require regulation. 
    To realize this, we identify positions of feature maps vital to previous tasks by $\mathscr{l}_1$-norm \cite{filterspruning} with a ratio $R$. 
	
	\subsubsection{Detailed Process}
	After training the \textsl{(t-1)}-th task, neural network $\theta^{t-1}$ is generated. We get important positions of feature maps from the first batch of \textsl{(t-1)}-th task. Assume position $(i,j)$ denotes \textsl{j}-th feature map in \text{i}-th layer. All important positions are recorded in a set $I$. Before training new \textsl{t}-th task, we first select a few samples, defined as $\mathcal M$, from the new task to run it on the neural network $\theta^{t-1}$ and get feature map standards, called $FS$, as references. The process is defined as
	$FS = F(\theta^{t-1}, \mathcal M, I) $,
    where $F$ is the function to get feature maps of important position $I$. Note that we only get $FS$ before training the new task without continually acquiring it during training, shown in the bottom of \figref{fig:pipeline}.
	
    During training the \textsl{t}-th task, we integrate $M$ in each batch and get the corresponding feature maps. $FM$ is calculated from $FM = F(\theta^{t}, \mathcal M, I)$. The limiting changes of vital feature maps between $FS$ and $FM$ can be described as the regulation loss function
	\begin{equation}
		\begin{gathered}[t]
			\mathcal L_{f} (FM, FS) = \beta \sum_{(i,j)\in I }(FM_{i,j}-FS_{i,j})^2.
		\end{gathered}
		\label{equ:regulate_1}
	\end{equation}
    The entire loss function $\mathcal{L}$ is the combination of cross-entropy loss function $\mathcal{L}_c$ and the regulation loss $\mathcal{L}_f$, described as
    \begin{equation}
		\begin{gathered}[t]
			\mathcal{L}(\theta^t) = \mathcal L_c(f(\theta^t, x_t), y_t) + \mathcal L_{f}(FM, FS).
		\end{gathered}
		\label{equ:regulate_2}
    \end{equation}
    The backward process of $\mathcal{L}$ is described as
    \begin{equation}
		\label{equ:regulate_3}
		\begin{array}{cl}
			\theta_{i,j}^t \leftarrow \theta_{i,j}^t + \alpha \nabla_{\theta_{i,j}^t} \mathcal L(f(\theta^t)) \quad
			\text{s.t.} \quad i > FZ.
		\end{array}
    \end{equation}
    
    After the \textsl{t}-th task, we use the first batch to get the important position $I'$ of the current task with a ratio $R$ and combine it with the last version, $I = I \lor I'$. The whole process of the algorithm is shown in \hyperref[Algorithm1]{Algorithm 1}.
	As shown in \secref{sec:Abaltion_regulate}, \textsl{Memorize Feature Patterns} successfully delays forgetting with a few extra resources. The combination of \textsl{Maintain Generalizability} and \textsl{Memorize Feature Patterns} outperforms other \sArt~ \CL~ methods, shown in the \tabref{Table1}.
	Different from other replay methods, we do not store samples in storage as they are only utilized during the current training. This also enhances privacy security.
    
	\IncMargin{1em}
	\begin{algorithm}[h]
		\SetKwData{Left}{left}\SetKwData{This}{this}\SetKwData{Up}{up} \SetKwFunction{Union}{Union}\SetKwFunction{FindCompress}{FindCompress} \SetKwInOut{Input}{Input}\SetKwInOut{Output}{output}\SetKwInOut{Initialize}{Initialize}
		\caption{\ourM~ Algorithm}
		\label{Algorithm1}
		\Input {Important Position Set $I$; Mem Sample $\mathcal M$; 
            Feature Standard $FS$; Feature Map $FM$; 
            Function of Getting Feature Map $F$; Function of Neural Network $f$; Cross-entropy loss $\mathcal L_c$; Feature Regularization $\mathcal L_{f}$; Selecting ratio $R$}
		\BlankLine 
        Freezing lower layers of a pre-trained neural network $\theta$ \\
		\For{$t=1,...,T$}{ 
			$\mathcal M \leftarrow sample(\mathcal D_t)$\\
			$FS \leftarrow F(\theta, \mathcal M, I)$
			
			\For{iteration $Batch$ from $\mathcal D_t$}{
				$(x,y) \leftarrow Batch$\\
				$FM \leftarrow F(\theta, \mathcal M, I)$\\
				$\mathcal L \leftarrow \mathcal L_c(f(\theta, x),y) + \mathcal L_{f}(FM, FS)$\\
				$\theta \leftarrow Update(\mathcal L, \theta)$
			}
			$(x,y) \leftarrow $ last $Batch$ from $\mathcal D_t$\\
			$FM \leftarrow F(\theta, x)$\\
			Select Important Position $I'$ with ratio $R$ from FM\\
			$I = I' \lor I$\\
		}
	\end{algorithm}
	\DecMargin{1em}

	\section{Experiment}\label{sec:experiment}
	
	\begin{table*}[ht]
		\centering
		\caption{Comparison with other \CL~ methods on pre-trained model. Classification results on ResNet-18 for standard \CL~ benchmarks, averaged across $5$ runs. \ourM~ is outstanding beyond other \CL~ methods with fewer resources. }
		\label{Table1} 
		\begin{tabular}{l|cc|cccc|cccc}
			\toprule	
			\multirow{3}{*}[-2pt]{\textbf{Method}}&
			\multirow{2}{*}{\textbf{Buffer}}&
			\multirow{3}{*}[-2pt]{\textbf{Sparsity}}&
			\multicolumn{4}{c|}{\textbf{Split CIFAR-10}}&
			\multicolumn{4}{c}{\textbf{Split TinyImageNet}}\\[2pt]&&&
			\multirow{2}{*}{CIL$(\uparrow)$}&
			\multirow{2}{*}{TIL$(\uparrow)$}&
			FLOPs&
			Mem&
			\multirow{2}{*}{CIL$(\uparrow)$}&
			\multirow{2}{*}{TIL$(\uparrow)$}&
			FLOPs&
			Mem\\&
			\textbf{size}&&&&
			$\times10^{15}(\downarrow)$&
			(MB)$(\downarrow)$&&&
			$\times10^{16}(\downarrow$)&
			(MB)$(\downarrow)$\\
			\midrule
			JOINT & \multirow{2}{*}[1pt]{/} & \multirow{2}{*}[1pt]{0.0} & 94.67\footnotesize{$\pm$0.15} & 99.04\footnotesize{$\pm$0.07} & 37.5 & 153.4 & 61.38\footnotesize{$\pm$0.12} & 83.14\footnotesize{$\pm$0.52}& 120 & 357.9\\
			SGD &&& 34.21\footnotesize{$\pm$3.06} & 86.19\footnotesize{$\pm$1.32} & 7.5 & 153.4 & 10.63\footnotesize{$\pm$0.36}& 34.24\footnotesize{$\pm$0.84} & 12 & 357.9\\
			\midrule
			EWC \cite{EWC} & \multirow{2}{*}{/} & \multirow{2}{*}{0.0} & 40.42\footnotesize{$\pm$3.81} & 89.77\footnotesize{$\pm$3.01} & 7.5 & 196.0 & 11.17\footnotesize{$\pm$1.55} & 34.70\footnotesize{$\pm$1.55} & 12 & 400.5\\
			PNN \cite{PNN} &&& / & 94.62\footnotesize{$\pm$0.52} & 10.5 & 323.9 & / & 73.94\footnotesize{$\pm$3.56} & 26.4 & 528.4\\
			\midrule
			FDR \cite{FDR} & \multirow{3}{*}[-2pt]{15} & \multirow{3}{*}[-2pt]{0.0} & 25.64\footnotesize{$\pm$2.52} & 85.54\footnotesize{$\pm$3.08} & 11 & 185.4 & 10.80\footnotesize{$\pm$0.71} & 34.42\footnotesize{$\pm$0.38} & 17.7 & 485.7\\
			ER \cite{er} &&&  54.44\footnotesize{$\pm$2.82} & 92.25\footnotesize{$\pm$1.92} & 11 &	185.4 & 11.82\footnotesize{$\pm$0.31} & 36.76\footnotesize{$\pm$0.57} & 17.7 & 485.7\\
			DER++ \cite{DER} &&& 40.75\footnotesize{$\pm$2.55} & 85.68\footnotesize{$\pm$4.35} & 14.5 & 217.3 & 7.98\footnotesize{$\pm$1.84} & 35.88\footnotesize{$\pm$0.41} & 23.3 & 613.5\\
			\midrule
			SparCL{\tiny{ER}} \cite{SparCL} & \multirow{2}{*}{15} & \multirow{2}{*}{0.9} &  36.98\footnotesize{$\pm$2.42} & 85.33\footnotesize{$\pm$2.49} & 1.1 &	106.6 & 10.01\footnotesize{$\pm$0.55} & 33.01\footnotesize{$\pm$0.78} & 1.8 & 407.0\\
			SparCL{\tiny{DER++}} \cite{SparCL} &&& 35.58\footnotesize{$\pm$0.84} & 85.28\footnotesize{$\pm$1.00} & 1.5 & 140.6 & 9.12\footnotesize{$\pm$1.02} & 31.86\footnotesize{$\pm$1.77} & 2.3 & 536.8\\
			\midrule
			\multirow{2}{*}{\textbf{\ourM}} & \multirow{2}{*}{\textbf{15}} & \textbf{0.0} & \textbf{55.53\footnotesize{$\pm$0.35}} & \textbf{96.28\footnotesize{$\pm$1.08}} & \textbf{5.4} & \textbf{92.8} & \textbf{12.79\footnotesize{$\pm$0.30}} & \textbf{37.39\footnotesize{$\pm$0.63}} & \textbf{8.6} & \textbf{128.0}\\ 
             && \textbf{0.9} & \textbf{38.31\footnotesize{$\pm$3.93}} & \textbf{86.88\footnotesize{$\pm$1.69}} & \textbf{2.3} & \textbf{25.0} & \textbf{8.95\footnotesize{$\pm$0.35}} & \textbf{33.94\footnotesize{$\pm$0.52}} & \textbf{3.7} & \textbf{87.1}\\
			\bottomrule
			\multicolumn{11}{l}{‘/’ indicates the corresponding item is not needed (Buffer is not required in JOINT, SGD, EWC, and PNN) or cannot run (PNN cannot run}\\
            \multicolumn{11}{l}{under CIL settings). FLOPs and Mem are calculated theoretically with the precision to one decimal place.}\\
		\end{tabular}
	\end{table*}

\begin{table}[ht]
		\centering
		\caption{Comparison with \CL~ methods from scratch with ResNet-18 and ResNet-50. \ourM~ still outperforms other \CL~ methods and shows outstanding performance on both large and small networks from scratch. }
		\label{Table2} 
		\begin{tabular}{l|cc|cc}
			\toprule	
			\multirow{2}{*}{\textbf{Method}}&
			\multicolumn{2}{c}{\textbf{Split CIFAR-10}}&
			\multicolumn{2}{c}{\textbf{Split TinyImageNet}}\\[2pt]&
			CIL$(\uparrow)$&
			TIL$(\uparrow)$&
			CIL$(\uparrow)$&
			TIL$(\uparrow)$\\
			\midrule
            \multicolumn{5}{c}{ResNet-18 }\\
            \midrule
			JOINT &92.20\footnotesize{$\pm$0.15} & 98.31\footnotesize{$\pm$0.12} & 59.99\footnotesize{$\pm$0.19} & 82.04\footnotesize{$\pm$0.10}\\
			SGD   &21.00\footnotesize{$\pm$1.61} & 65.78\footnotesize{$\pm$3.52} & 7.28\footnotesize{$\pm$0.04} & 24.99\footnotesize{$\pm$0.27}\\
			\midrule
			EWC   &21.32\footnotesize{$\pm$0.17} & 66.13\footnotesize{$\pm$0.17} & 8.58\footnotesize{$\pm$0.22} & 24.47\footnotesize{$\pm$0.30}\\
			PNN   &   /                          & 95.13\footnotesize{$\pm$0.72} & / & 61.88\footnotesize{$\pm$1.00}\\
			\midrule
			FDR   &26.19\footnotesize{$\pm$3.14} & 77.61\footnotesize{$\pm$0.68} & 8.58 \footnotesize{$\pm$0.21} & 25.20\footnotesize{$\pm$0.29}\\
			ER    &35.90\footnotesize{$\pm$1.53} & 82.40\footnotesize{$\pm$1.64} & 8.76\footnotesize{$\pm$0.23}  & 26.29\footnotesize{$\pm$1.18}\\
			DER++ &36.25\footnotesize{$\pm$1.99} & 82.05\footnotesize{$\pm$1.63} & 6.53\footnotesize{$\pm$1.44}  & 27.21\footnotesize{$\pm$1.85}\\
			\midrule
			SparCL{\tiny{ER}} &30.09\footnotesize{$\pm$1.60} & 75.12\footnotesize{$\pm$0.63} & 7.70\footnotesize{$\pm$0.13} & 25.24\footnotesize{$\pm$0.55}\\
			SparCL{\tiny{DER++}} &34.36\footnotesize{$\pm$1.32} & 78.56\footnotesize{$\pm$2.88}&8.16\footnotesize{$\pm$0.58} & 26.53\footnotesize{$\pm$0.54}\\
			\midrule
			\textbf{\ourM}        &\textbf{37.86\footnotesize{$\pm$2.16}} & \textbf{85.69\footnotesize{$\pm$0.98}} & 
            \textbf{9.03\footnotesize{$\pm$0.16}} &
            \textbf{33.74\textbf{\footnotesize{$\pm$0.68}}} \\
            \textbf{\ourM$_{0.9}$}&\textbf{35.72\footnotesize{$\pm$2.67}} & \textbf{83.37\footnotesize{$\pm$1.10}} &
            \textbf{8.76\footnotesize{$\pm$0.63}} &
            \textbf{36.86\textbf{\footnotesize{$\pm$0.29}}}\\
            \midrule
            \multicolumn{5}{c}{ResNet-50}\\
            \midrule
            JOINT & 94.66\footnotesize{$\pm$0.15} & 98.93\footnotesize{$\pm$0.09} & 61.04\footnotesize{$\pm$0.42} & 82.70\footnotesize{$\pm$0.42}\\
			SGD   &14.16\footnotesize{$\pm$0.39} & 72.05\footnotesize{$\pm$1.52} & 8.55\footnotesize{$\pm$0.20} & 29.99\footnotesize{$\pm$0.23}\\
			\midrule
			ER    &24.19\footnotesize{$\pm$1.43} & 76.65\footnotesize{$\pm$1.50} & 8.81\footnotesize{$\pm$0.62}  & 31.38\footnotesize{$\pm$0.06}\\
			DER++ &26.42\footnotesize{$\pm$1.67} & 75.13\footnotesize{$\pm$0.69} & 6.23\footnotesize{$\pm$1.23}  & 27.22\footnotesize{$\pm$2.70}\\
			\midrule
			\textbf{\ourM}        &\textbf{35.08\footnotesize{$\pm$0.50}} & \textbf{81.49\footnotesize{$\pm$0.92}} & 
            \textbf{9.23\footnotesize{$\pm$0.67}} &
            \textbf{33.81\textbf{\footnotesize{$\pm$1.47}}} \\
            \midrule
			\bottomrule
		\end{tabular}
	\end{table}

	\subsection{Experiment Setting}\label{sec:Experiment_setting}
	\noindent \textbf{Datasets.} We evaluate \ourM~ on Split CIFAR-10 \cite{Cifar10} and Split Tiny-ImageNet \cite{Imagenet}. To compare with other \CL~ methods, we strictly follow \cite{DER, SparCL} by splitting CIFAR-10 and Tiny-ImageNet into $5$ and $10$ tasks, and each task including $2$ and $20$ classes, respectively. For Split CIFAR-10 and Tiny-ImageNet, we train each task for $50$ and $100$ epochs with a batch size of $32$.
	
	\noindent \textbf{Baseline Mehtods.} We select typical \CL~ methods from three categories mentioned in section \ref{sec:relate}, including regularization-based (EWC \cite{EWC}), replay (DER \cite{DER}, ER \cite{er}, and FDR \cite{FDR}), and architecture-based (PNN \cite{PNN}) methods. Since these above methods do not consider the edge scenario, we also compare our method against the \sArt~ efficient \CL~ method SparCL \cite{SparCL} combined with the replay method DER++ and ER. We use SGD without any \CL~ method as a lower bound and train all tasks jointly (JOINT) as an upper bound.
	
	\noindent \textbf{Metrics.} We use AA, mentioned in \secref{sec:CL-setting}, to evaluate the performance of overcoming forgetting. Moreover, we follow \cite{SparCL} to theoretically assess the training FLOPs and peak memory footprint (including feature maps, gradients, and model parameters during training) for the illustration of resource usage.

    \noindent \textbf{Setting of \ourM.} We use a pre-trained ResNet-18 model (pre-trained on the ImageNet$32$$\times$$32$ \cite{Image32}), and freeze all lower and middle layers (before Layer4.0.conv1 in ResNet-18). For regulating the feature maps, we select vital feature maps with a ratio $R=15$\% in each layer after Layer4.0.conv1 (included). Hyperparameter $\beta$ in Equation \ref{equ:regulate_1} equals $0.0002$. Although our method does not need to store samples in storage, we set the same buffer size to compare \ourM~ with replay methods for a fair comparison. 
	
    \noindent \textbf{Experiment Details.} We operate our experiments on a server equipped with an NVIDIA A800 GPU in \secref{sec:Main_Result} and \secref{sec:Ablation}.
    We further verify the practical effectiveness of our method on an edge device, Jetson Nano, in \secref{sec:Deployment on Edge Device}. All methods employ ResNet-18 \cite{resnet} with a learning rate of $0.01$. We set other hyperparameters as suggested in \cite{DER,SparCL} for other \CL~ methods. 
    For a fair comparison, all other methods employ the same pre-trained model in \secref{sec:compare_pretrain}. Although the pre-trained model is more practical for application, most of \CL~ methods do not take this into account and pay more attention to training the model from scratch. Hence, we test all methods without the pre-trained model to evaluate the effectiveness of \ourM~ in \secref{sec:from scratch}. In order to compare with SparCL, we prune our model with sparsity ratio $90\%$ as the same as SparCL. Specifically, \ourM~ uses static filter pruning for ResNet-18 and sets sparsity ratio as follows: 1) before Layer3.0 as $20\%$; 2) Layer3.0 and Layer3.1 as $70\%$; 3) others as $80\%$. 
    Sparse is not the key point in our paper, and we set it up this way because we want to retain enough knowledge in the frozen layers.
	
	\subsection{Main Result}\label{sec:Main_Result}
	\subsubsection{Comparison with \CL~ methods on the pre-trained model.}\label{sec:compare_pretrain}
    We evaluate results on Split CIFAR-10 and Tiny-ImageNet with a pre-trained ResNet-18 model under both CIL and TIL settings. The compared methods are divided into normal \CL~ methods (EWC, PNN, FDR, ER, and DER++) and an efficient \CL~ method (SparCL). As shown in \tabref{Table1},
    the accuracy of \ourM~ without sparsity is higher than ER, which performs best in normal \CL~ methods except PNN. On Split Tiny-ImageNet, FLOPs of \ourM~ greatly reduce at most $3.07$$\times$ and Memory footprint at most $4.79$$\times$, compared with normal \CL~ methods. That is because \ourM~ reduces the computation of backward process, and memory footprint of gradient and intermediate feature maps in frozen layers. However, other algorithms significantly increase resource consumption to mitigate forgetting. For example, EWC requires storing twice the weight size because it calculates the importance of each weight. In the Replay method (FDR, ER, and DER++), during training a new task, previous samples from the buffer need to be included in each batch, which greatly increases memory and computation consumption. Among them, DER++ is particularly severe as it needs to calculate twice the buffer size, leading to even greater increases. 
    Although the architecture-based method PNN has excellent accuracy under the TIL setting, it has tremendous FLOPs and memory footprint usage since it requires an additional structure for a new task and its forward process requires all structures during training. It also ignores the real application always under the CIL setting where the identity of a certain task is not known during testing.
    On the other hand, compared with the efficient method SparCL, the delaying performance of \ourM~ with the same sparsity ratio is still competitive. 
    Note that we employ a simple sparsity strategy without any optimization, such as dynamic sparsity \cite{SparCL}. 
    Surprisingly, the memory footprint is reduced at most $6.19$$\times$ compared with SparCL$_{DER++}$.
    Although the FLOPs of \ourM~ is higher than SparCL, \ourM~ is faster than SparCL in the real application since SparCL still costs abundant time on memory transfer. This will be discussed in \secref{sec:Deployment on Edge Device}. 
        
	\subsubsection{Comparison from scratch.} \label{sec:from scratch}
	Although deploying a pre-trained model on edge devices is more practical, most \CL~ methods ignore it and focus more on \CL~ from scratch. To better illustrate the advantages of \ourM, we also compare \ourM~ with other \CL~ methods in the setting of \CL~ from scratch. To further demonstrate the performance of \ourM~ on a large network, we compare \ourM~ with outstanding CL methods (ER and DER++) on ResNet-50. Since \ourM~ needs to freeze the lower and middle layers, we take the first task in the dataset as a pre-trained process, and we freeze the lower and middle layers in the following tasks. For simplicity, the experimental settings of ResNet-50 are the same as ResNet-18 and we set frozen layers before layer4.0.conv1 in ResNet-50. From \tabref{Table2}, \ourM~ is still outstanding compared with other methods in both ResNet-18 and ResNet-50. When using ResNet-18, \ourM$_{0.9}$ with sparsity ratio $90\%$ also shows better performance compared with the SparCL. Although replay methods (FDR, ER, and DER++) are ineffective with a limited buffer size of $15$, \ourM~ is favorable in a limited buffer size.

	\subsection{Ablation}\label{sec:Ablation}
\begin{figure}[ht]
            \centering
            \includegraphics[scale=0.06]{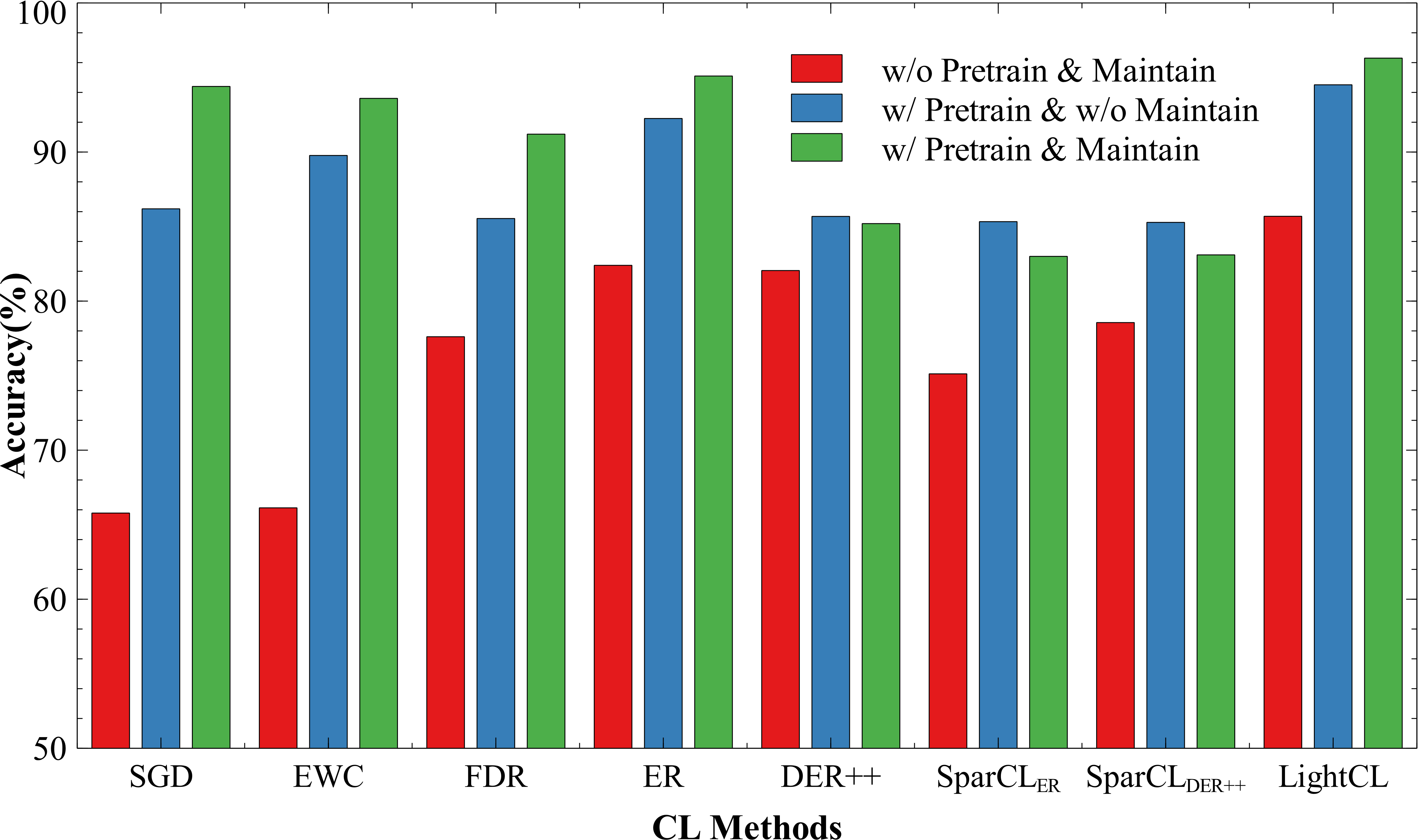}
            \caption{Ablation experiments on Split CIFAR-10 dataset under the TIL setting across 5 runs. Red bars denote the setting without the pre-trained model and \textsl{Maintain Generalizability} (freezing layers). Blue bars denote the setting only with the pre-trained model. Green bars have both of them.}
            \label{fig:maintain}
        \end{figure}

        In this section, we illustrate the effectiveness of \textsl{Maintain Generalizability} and \textsl{Memorize Feature Patterns} of \ourM~ with ablation experiments in \figref{fig:maintain}.
        
	\noindent \textbf{Discussion of Maintain Generalizability. }\label{sec:Ablation_maintain}
        \textsl{Maintain Generalizability} is an easy but effective method to delay forgetting and reduce resource usage. From \figref{fig:maintain}, compared with red and blue bars, all \CL~ methods can enhance performance with the pre-trained model. Most of them (SGD, EWC, FDR, ER, and \ourM) can even further delay the forgetting problem by \textsl{Maintain Generalizability}, shown in green bars. Note that the accuracy of SparCL decreases when using \textsl{Maintain Generalizability}. This phenomenon occurs because frozen layers weaken the dynamic pruning process, leading to a drop in accuracy. Although most of \CL~ methods can benefit from \textsl{Maintain Generalizability}, \ourM~ still performs better than other methods.

        \noindent \textbf{Discussion of Memorize Feature Patterns. }\label{sec:Abaltion_regulate}
        We compare the performance of \ourM~ with and without \textsl{Memorize Feature Patterns}.
        As shown in \figref{fig:maintain}, the green bars of SGD and \ourM~ are able to represent the \ourM~ with and without \textsl{Memorize Feature Patterns}, respectively. The result shows that \ourM~ with \textsl{Memorize Feature Patterns} can further increase the accuracy, proving its effectiveness. Considering the resource consumption of this method, we need to store the Feature Standard $FS$, mentioned in \secref{sec:Memorize Feature Patterns}, as a reference to regulate. As we only need to store important feature maps after Layer4.0.conv1 (included), the overall consumption of memory footprint, suppose all feature maps need to be stored, occupies about $1.9$MB (on Split CIFAR-10), which is pretty small compared to what we reduce by \textsl{Maintain Generalizability}.
        
	\subsection{Verification on Edge Device}\label{sec:Deployment on Edge Device}
\begin{figure}[ht]
            \centering
            \includegraphics[scale=0.1]{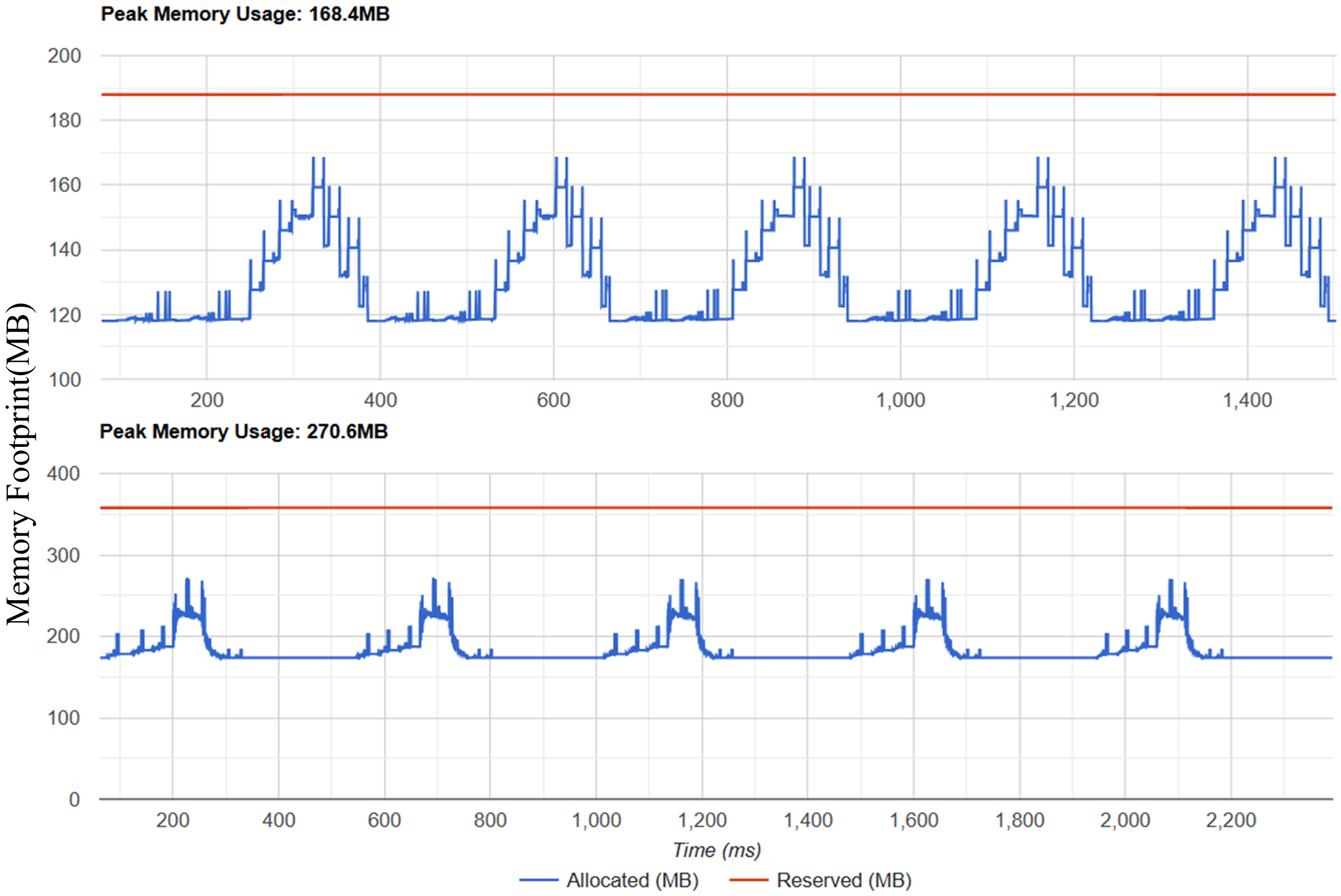}
            \caption{We deploy LightCL (top figure) and SparCL$_{DER++}$ (bottom figure) on Jetson Nano and evaluate the dynamic change of memory footprint during CL. Only $5$ iterations are selected to illustrate this experiment.}
            \label{fig:on_device}
        \end{figure}

 We deploy \ourM~ on an edge device, Jetson Nano, to verify the efficiency in a realistic environment. Although we theoretically calculate the peak memory footprint for model parameters, gradients, and intermediate feature maps, some real-applied consumption, such as extra consumption in loading datasets, memory fragmentation, and some additional intermediate variables, also need to be considered. Hence, we conduct an experiment to compare \ourM~ with SparCL$_{DER++}$ on an edge device. As shown in \figref{fig:on_device}, although the peak memory of them is all greater than the theoretical value shown in \tabref{Table1}, the peak memory of LightCL is still greatly smaller compared to SparCL. Note that the consumption of the memory footprint is not always equal to the peak memory footprint. We also need to observe the reserved memory, red line in \figref{fig:on_device}. Since the peak memory of \ourM~ is smaller than SparCL, the reserved memory is also smaller than SparCL. From the x-axis that represents the change in time, the training process of \ourM~ is faster than SparCL because the memory transfer consumption has greatly hindered the training speed in SparCL. These findings collectively demonstrate the effectiveness of LightCL on edge devices. 
\section{Conclusion}\label{sec:conclusion}
	This paper proposes a compact \CL~ algorithm \ourM~ to overcome the catastrophic forgetting problem in edge scenarios. It is the first study to explore the distribution of generalizability in neural networks by evaluating two designed metrics of learning plasticity and memory stability. This evaluation demonstrates that the generalizability of different layers in a neural network exhibits a significant variation. Thus, we propose \ourM~ with \textsl{Maintaining Generalizability} and \textsl{Memorize Feature Patterns}.
    Experiments demonstrate that \ourM~ outpeforms other \sArt~ \CL~ methods with fewer resources, especially a lower memory footprint. Also, our method performs more favorably on edge devices and shows significant potential for efficient CL.

\section{Acknowledgments}
This work was supported by Educational Teaching Reform Practice Program No.A-XSY2308,  University-Industry Collaborative Education Program No.231103428160213, Proof of Concept Foundation of Xidian University Hangzhou Institute of Technology under Grant No.GNYZ2023GY0301, and the Fundamental Research Funds for the Central Universities No.QTZX24064. Thanks to the help provided by the National Experimental Teaching Demonstration Center for Computer Network and Information Security affiliated with Xidian University. Also, we are grateful to Qichang Zhang for checking typos and grammar errors.
\bibliographystyle{ACM-Reference-Format}
\bibliography{reference}


\begin{thebibliography}{31}

%%% the Web address.

\ifx \showCODEN    \undefined \def \showCODEN     #1{\unskip}     \fi
\ifx \showDOI      \undefined \def \showDOI       #1{#1}\fi
\ifx \showISBNx    \undefined \def \showISBNx     #1{\unskip}     \fi
\ifx \showISBNxiii \undefined \def \showISBNxiii  #1{\unskip}     \fi
\ifx \showISSN     \undefined \def \showISSN      #1{\unskip}     \fi
\ifx \showLCCN     \undefined \def \showLCCN      #1{\unskip}     \fi
\ifx \shownote     \undefined \def \shownote      #1{#1}          \fi
\ifx \showarticletitle \undefined \def \showarticletitle #1{#1}   \fi
\ifx \showURL      \undefined \def \showURL       {\relax}        \fi
% invisible to TeX
\providecommand\bibfield[2]{#2}
\providecommand\bibinfo[2]{#2}
\providecommand\natexlab[1]{#1}
\providecommand\showeprint[2][]{arXiv:#2}

\bibitem[Aljundi et~al\mbox{.}(2018)]%
        {MAS}
\bibfield{author}{\bibinfo{person}{Rahaf Aljundi}, \bibinfo{person}{Francesca Babiloni}, \bibinfo{person}{Mohamed Elhoseiny}, \bibinfo{person}{Marcus Rohrbach}, {and} \bibinfo{person}{Tinne Tuytelaars}.} \bibinfo{year}{2018}\natexlab{}.
\newblock \showarticletitle{Memory Aware Synapses: Learning what (not) to forget}. In \bibinfo{booktitle}{\emph{Proceedings of the European Conference on Computer Vision (ECCV)}}.
\newblock


\bibitem[Benjamin et~al\mbox{.}(2019)]%
        {FDR}
\bibfield{author}{\bibinfo{person}{Ari Benjamin}, \bibinfo{person}{David Rolnick}, {and} \bibinfo{person}{Konrad Kording}.} \bibinfo{year}{2019}\natexlab{}.
\newblock \showarticletitle{Measuring and regularizing networks in function space}. In \bibinfo{booktitle}{\emph{International Conference on Learning Representations}}.
\newblock


\bibitem[Buzzega et~al\mbox{.}(2020)]%
        {DER}
\bibfield{author}{\bibinfo{person}{Pietro Buzzega}, \bibinfo{person}{Matteo Boschini}, \bibinfo{person}{Angelo Porrello}, \bibinfo{person}{Davide Abati}, {and} \bibinfo{person}{SIMONE CALDERARA}.} \bibinfo{year}{2020}\natexlab{}.
\newblock \showarticletitle{Dark Experience for General Continual Learning: a Strong, Simple Baseline}. In \bibinfo{booktitle}{\emph{Advances in Neural Information Processing Systems}}, Vol.~\bibinfo{volume}{33}. \bibinfo{pages}{15920--15930}.
\newblock


\bibitem[Cha et~al\mbox{.}(2021)]%
        {Co2l}
\bibfield{author}{\bibinfo{person}{Hyuntak Cha}, \bibinfo{person}{Jaeho Lee}, {and} \bibinfo{person}{Jinwoo Shin}.} \bibinfo{year}{2021}\natexlab{}.
\newblock \showarticletitle{Co2L: Contrastive Continual Learning}. In \bibinfo{booktitle}{\emph{2021 IEEE/CVF International Conference on Computer Vision (ICCV)}}. \bibinfo{pages}{9496--9505}.
\newblock


\bibitem[Chaudhry et~al\mbox{.}(2019)]%
        {er}
\bibfield{author}{\bibinfo{person}{Arslan Chaudhry}, \bibinfo{person}{Marcus Rohrbach}, \bibinfo{person}{Mohamed Elhoseiny}, \bibinfo{person}{Thalaiyasingam Ajanthan}, \bibinfo{person}{Puneet~Kumar Dokania}, \bibinfo{person}{Philip H.~S. Torr}, {and} \bibinfo{person}{Marc'Aurelio Ranzato}.} \bibinfo{year}{2019}\natexlab{}.
\newblock \showarticletitle{Continual Learning with Tiny Episodic Memories}.
\newblock \bibinfo{journal}{\emph{CoRR}}  \bibinfo{volume}{abs/1902.10486} (\bibinfo{year}{2019}).
\newblock


\bibitem[Chrabaszcz et~al\mbox{.}(2017)]%
        {Image32}
\bibfield{author}{\bibinfo{person}{Patryk Chrabaszcz}, \bibinfo{person}{Ilya Loshchilov}, {and} \bibinfo{person}{Frank Hutter}.} \bibinfo{year}{2017}\natexlab{}.
\newblock \bibinfo{title}{A Downsampled Variant of ImageNet as an Alternative to the CIFAR datasets}.
\newblock
\newblock
\showeprint[arxiv]{1707.08819}~[cs.CV]


\bibitem[Deng et~al\mbox{.}(2009)]%
        {Imagenet}
\bibfield{author}{\bibinfo{person}{Jia Deng}, \bibinfo{person}{Wei Dong}, \bibinfo{person}{Richard Socher}, \bibinfo{person}{Li-Jia Li}, \bibinfo{person}{Kai Li}, {and} \bibinfo{person}{Li Fei-Fei}.} \bibinfo{year}{2009}\natexlab{}.
\newblock \showarticletitle{ImageNet: A large-scale hierarchical image database}. In \bibinfo{booktitle}{\emph{2009 IEEE Conference on Computer Vision and Pattern Recognition}}. \bibinfo{pages}{248--255}.
\newblock


\bibitem[Gholami et~al\mbox{.}(2024)]%
        {memorywall}
\bibfield{author}{\bibinfo{person}{Amir Gholami}, \bibinfo{person}{Zhewei Yao}, \bibinfo{person}{Sehoon Kim}, \bibinfo{person}{Coleman Hooper}, \bibinfo{person}{Michael~W. Mahoney}, {and} \bibinfo{person}{Kurt Keutzer}.} \bibinfo{year}{2024}\natexlab{}.
\newblock \showarticletitle{AI and Memory Wall}.
\newblock \bibinfo{journal}{\emph{IEEE Micro}} \bibinfo{volume}{44}, \bibinfo{number}{3} (\bibinfo{date}{May} \bibinfo{year}{2024}), \bibinfo{pages}{33--39}.
\newblock
\showISSN{1937-4143}


\bibitem[Gong et~al\mbox{.}(2014)]%
        {Quantization}
\bibfield{author}{\bibinfo{person}{Yunchao Gong}, \bibinfo{person}{Liu Liu}, \bibinfo{person}{Ming Yang}, {and} \bibinfo{person}{Lubomir~D. Bourdev}.} \bibinfo{year}{2014}\natexlab{}.
\newblock \showarticletitle{Compressing Deep Convolutional Networks using Vector Quantization}.
\newblock \bibinfo{journal}{\emph{CoRR}}  \bibinfo{volume}{abs/1412.6115} (\bibinfo{year}{2014}).
\newblock
\showeprint[arXiv]{1412.6115}


\bibitem[Han et~al\mbox{.}(2016)]%
        {deepcompression}
\bibfield{author}{\bibinfo{person}{Song Han}, \bibinfo{person}{Huizi Mao}, {and} \bibinfo{person}{William~J Dally}.} \bibinfo{year}{2016}\natexlab{}.
\newblock \showarticletitle{Deep compression: Compressing deep neural networks with pruning, trained quantization and huffman coding}.
\newblock \bibinfo{journal}{\emph{ICLR}} (\bibinfo{year}{2016}).
\newblock


\bibitem[Han et~al\mbox{.}(2015)]%
        {prune-retrain}
\bibfield{author}{\bibinfo{person}{Song Han}, \bibinfo{person}{Jeff Pool}, \bibinfo{person}{John Tran}, {and} \bibinfo{person}{William Dally}.} \bibinfo{year}{2015}\natexlab{}.
\newblock \showarticletitle{Learning both Weights and Connections for Efficient Neural Network}. In \bibinfo{booktitle}{\emph{Advances in Neural Information Processing Systems}}, Vol.~\bibinfo{volume}{28}.
\newblock


\bibitem[Hayes et~al\mbox{.}(2020)]%
        {remind}
\bibfield{author}{\bibinfo{person}{Tyler~L. Hayes}, \bibinfo{person}{Kushal Kafle}, \bibinfo{person}{Robik Shrestha}, \bibinfo{person}{Manoj Acharya}, {and} \bibinfo{person}{Christopher Kanan}.} \bibinfo{year}{2020}\natexlab{}.
\newblock \showarticletitle{REMIND Your Neural Network to Prevent Catastrophic Forgetting}. In \bibinfo{booktitle}{\emph{Computer Vision -- ECCV 2020}}. \bibinfo{publisher}{Springer International Publishing}, \bibinfo{address}{Cham}, \bibinfo{pages}{466--483}.
\newblock
\showISBNx{978-3-030-58598-3}


\bibitem[He et~al\mbox{.}(2016)]%
        {resnet}
\bibfield{author}{\bibinfo{person}{Kaiming He}, \bibinfo{person}{Xiangyu Zhang}, \bibinfo{person}{Shaoqing Ren}, {and} \bibinfo{person}{Jian Sun}.} \bibinfo{year}{2016}\natexlab{}.
\newblock \showarticletitle{Deep Residual Learning for Image Recognition}. In \bibinfo{booktitle}{\emph{Proceedings of the IEEE Conference on Computer Vision and Pattern Recognition (CVPR)}}.
\newblock


\bibitem[Kirkpatrick et~al\mbox{.}(2017)]%
        {EWC}
\bibfield{author}{\bibinfo{person}{James Kirkpatrick}, \bibinfo{person}{Razvan Pascanu}, \bibinfo{person}{Neil Rabinowitz}, \bibinfo{person}{Joel Veness}, \bibinfo{person}{Guillaume Desjardins}, \bibinfo{person}{Andrei~A. Rusu}, \bibinfo{person}{Kieran Milan}, \bibinfo{person}{John Quan}, \bibinfo{person}{Tiago Ramalho}, \bibinfo{person}{Agnieszka Grabska-Barwinska}, \bibinfo{person}{Demis Hassabis}, \bibinfo{person}{Claudia Clopath}, \bibinfo{person}{Dharshan Kumaran}, {and} \bibinfo{person}{Raia Hadsell}.} \bibinfo{year}{2017}\natexlab{}.
\newblock \showarticletitle{Overcoming catastrophic forgetting in neural networks}.
\newblock \bibinfo{journal}{\emph{Proceedings of the National Academy of Sciences}} \bibinfo{volume}{114}, \bibinfo{number}{13} (\bibinfo{year}{2017}), \bibinfo{pages}{3521--3526}.
\newblock


\bibitem[Krizhevsky et~al\mbox{.}(2009)]%
        {Cifar10}
\bibfield{author}{\bibinfo{person}{Alex Krizhevsky}, \bibinfo{person}{Geoffrey Hinton}, {et~al\mbox{.}}} \bibinfo{year}{2009}\natexlab{}.
\newblock \showarticletitle{Learning multiple layers of features from tiny images}.
\newblock \bibinfo{journal}{\emph{https://www.cs.toronto.edu/~kriz/learning-features-2009-TR.pdf}} (\bibinfo{year}{2009}).
\newblock


\bibitem[Lee et~al\mbox{.}(2020)]%
        {cloud}
\bibfield{author}{\bibinfo{person}{Changha Lee}, \bibinfo{person}{Seong-Hwan Kim}, {and} \bibinfo{person}{Chan-Hyun Youn}.} \bibinfo{year}{2020}\natexlab{}.
\newblock \showarticletitle{An Accelerated Continual Learning with Demand Prediction based Scheduling in Edge-Cloud Computing}. In \bibinfo{booktitle}{\emph{2020 International Conference on Data Mining Workshops (ICDMW)}}. \bibinfo{pages}{717--722}.
\newblock


\bibitem[Li et~al\mbox{.}(2017)]%
        {filterspruning}
\bibfield{author}{\bibinfo{person}{Hao Li}, \bibinfo{person}{Asim Kadav}, \bibinfo{person}{Igor Durdanovic}, \bibinfo{person}{Hanan Samet}, {and} \bibinfo{person}{Hans~Peter Graf}.} \bibinfo{year}{2017}\natexlab{}.
\newblock \showarticletitle{Pruning Filters for Efficient ConvNets}. In \bibinfo{booktitle}{\emph{International Conference on Learning Representations}}.
\newblock


\bibitem[Li and Hoiem(2018)]%
        {LWF}
\bibfield{author}{\bibinfo{person}{Zhizhong Li} {and} \bibinfo{person}{Derek Hoiem}.} \bibinfo{year}{2018}\natexlab{}.
\newblock \showarticletitle{Learning without Forgetting}.
\newblock \bibinfo{journal}{\emph{IEEE Transactions on Pattern Analysis and Machine Intelligence}} \bibinfo{volume}{40}, \bibinfo{number}{12} (\bibinfo{year}{2018}), \bibinfo{pages}{2935--2947}.
\newblock


\bibitem[Lin et~al\mbox{.}(2022)]%
        {256kb}
\bibfield{author}{\bibinfo{person}{Ji Lin}, \bibinfo{person}{Ligeng Zhu}, \bibinfo{person}{Wei-Ming Chen}, \bibinfo{person}{Wei-Chen Wang}, \bibinfo{person}{Chuang Gan}, {and} \bibinfo{person}{Song Han}.} \bibinfo{year}{2022}\natexlab{}.
\newblock \showarticletitle{On-Device Training Under 256KB Memory}. In \bibinfo{booktitle}{\emph{Advances in Neural Information Processing Systems}}, Vol.~\bibinfo{volume}{35}. \bibinfo{pages}{22941--22954}.
\newblock


\bibitem[Lin et~al\mbox{.}(2023)]%
        {TinyML-survey}
\bibfield{author}{\bibinfo{person}{Ji Lin}, \bibinfo{person}{Ligeng Zhu}, \bibinfo{person}{Wei-Ming Chen}, \bibinfo{person}{Wei-Chen Wang}, {and} \bibinfo{person}{Song Han}.} \bibinfo{year}{2023}\natexlab{}.
\newblock \showarticletitle{Tiny Machine Learning: Progress and Futures [Feature]}.
\newblock \bibinfo{journal}{\emph{IEEE Circuits and Systems Magazine}} \bibinfo{volume}{23}, \bibinfo{number}{3} (\bibinfo{year}{2023}), \bibinfo{pages}{8--34}.
\newblock


\bibitem[Mallya and Lazebnik(2018)]%
        {Packnet}
\bibfield{author}{\bibinfo{person}{Arun Mallya} {and} \bibinfo{person}{Svetlana Lazebnik}.} \bibinfo{year}{2018}\natexlab{}.
\newblock \showarticletitle{PackNet: Adding Multiple Tasks to a Single Network by Iterative Pruning}. In \bibinfo{booktitle}{\emph{Proceedings of the IEEE Conference on Computer Vision and Pattern Recognition (CVPR)}}.
\newblock


\bibitem[McCloskey and Cohen(1989)]%
        {forgetting}
\bibfield{author}{\bibinfo{person}{Michael McCloskey} {and} \bibinfo{person}{{Neal J.} Cohen}.} \bibinfo{year}{1989}\natexlab{}.
\newblock \showarticletitle{Catastrophic Interference in Connectionist Networks: The Sequential Learning Problem}.
\newblock \bibinfo{journal}{\emph{Psychology of Learning and Motivation - Advances in Research and Theory}} \bibinfo{volume}{24}, \bibinfo{number}{C} (\bibinfo{date}{1 Jan.} \bibinfo{year}{1989}), \bibinfo{pages}{109--165}.
\newblock
\showISSN{0079-7421}


\bibitem[Ramasesh et~al\mbox{.}(2021)]%
        {anatomy}
\bibfield{author}{\bibinfo{person}{Vinay~Venkatesh Ramasesh}, \bibinfo{person}{Ethan Dyer}, {and} \bibinfo{person}{Maithra Raghu}.} \bibinfo{year}{2021}\natexlab{}.
\newblock \showarticletitle{Anatomy of Catastrophic Forgetting: Hidden Representations and Task Semantics}. In \bibinfo{booktitle}{\emph{International Conference on Learning Representations}}.
\newblock


\bibitem[Ren and Honavar(2024)]%
        {EsaCL}
\bibfield{author}{\bibinfo{person}{Weijieying Ren} {and} \bibinfo{person}{Vasant~G. Honavar}.} \bibinfo{year}{2024}\natexlab{}.
\newblock \showarticletitle{EsaCL: An Efficient Continual Learning Algorithm}. In \bibinfo{booktitle}{\emph{Proceedings of the 2024 {SIAM} International Conference on Data Mining, {SDM} 2024, Houston, TX, USA, April 18-20, 2024}}, \bibfield{editor}{\bibinfo{person}{Shashi Shekhar}, \bibinfo{person}{Vagelis Papalexakis}, \bibinfo{person}{Jing Gao}, \bibinfo{person}{Zhe Jiang}, {and} \bibinfo{person}{Matteo Riondato}} (Eds.). \bibinfo{publisher}{{SIAM}}, \bibinfo{pages}{163--171}.
\newblock


\bibitem[{Rusu} et~al\mbox{.}(2016)]%
        {PNN}
\bibfield{author}{\bibinfo{person}{Andrei~A. {Rusu}}, \bibinfo{person}{Neil~C. {Rabinowitz}}, \bibinfo{person}{Guillaume {Desjardins}}, \bibinfo{person}{Hubert {Soyer}}, \bibinfo{person}{James {Kirkpatrick}}, \bibinfo{person}{Koray {Kavukcuoglu}}, \bibinfo{person}{Razvan {Pascanu}}, {and} \bibinfo{person}{Raia {Hadsell}}.} \bibinfo{year}{2016}\natexlab{}.
\newblock \showarticletitle{{Progressive Neural Networks}}.
\newblock \bibinfo{journal}{\emph{arXiv e-prints}} (\bibinfo{date}{June} \bibinfo{year}{2016}).
\newblock
\showeprint[arxiv]{1606.04671}


\bibitem[Shuvo et~al\mbox{.}(2023)]%
        {inference}
\bibfield{author}{\bibinfo{person}{Md. Maruf~Hossain Shuvo}, \bibinfo{person}{Syed~Kamrul Islam}, \bibinfo{person}{Jianlin Cheng}, {and} \bibinfo{person}{Bashir~I. Morshed}.} \bibinfo{year}{2023}\natexlab{}.
\newblock \showarticletitle{Efficient Acceleration of Deep Learning Inference on Resource-Constrained Edge Devices: A Review}.
\newblock \bibinfo{journal}{\emph{Proc. IEEE}} \bibinfo{volume}{111}, \bibinfo{number}{1} (\bibinfo{year}{2023}), \bibinfo{pages}{42--91}.
\newblock


\bibitem[Simonyan and Zisserman(2015)]%
        {VGG}
\bibfield{author}{\bibinfo{person}{Karen Simonyan} {and} \bibinfo{person}{Andrew Zisserman}.} \bibinfo{year}{2015}\natexlab{}.
\newblock \showarticletitle{Very Deep Convolutional Networks for Large-Scale Image Recognition}. In \bibinfo{booktitle}{\emph{3rd International Conference on Learning Representations, {ICLR} 2015, San Diego, CA, USA, May 7-9, 2015, Conference Track Proceedings}}.
\newblock


\bibitem[Wang et~al\mbox{.}(2020)]%
        {LPS}
\bibfield{author}{\bibinfo{person}{Zifeng Wang}, \bibinfo{person}{Tong Jian}, \bibinfo{person}{Kaushik Chowdhury}, \bibinfo{person}{Yanzhi Wang}, \bibinfo{person}{Jennifer Dy}, {and} \bibinfo{person}{Stratis Ioannidis}.} \bibinfo{year}{2020}\natexlab{}.
\newblock \showarticletitle{Learn-Prune-Share for Lifelong Learning}. In \bibinfo{booktitle}{\emph{2020 IEEE International Conference on Data Mining (ICDM)}}. \bibinfo{pages}{641--650}.
\newblock


\bibitem[Wang et~al\mbox{.}(2023)]%
        {2023survey}
\bibfield{author}{\bibinfo{person}{Zhenyi Wang}, \bibinfo{person}{Enneng Yang}, \bibinfo{person}{Li Shen}, {and} \bibinfo{person}{Heng Huang}.} \bibinfo{year}{2023}\natexlab{}.
\newblock \showarticletitle{A Comprehensive Survey of Forgetting in Deep Learning Beyond Continual Learning}.
\newblock \bibinfo{journal}{\emph{CoRR}}  \bibinfo{volume}{abs/2307.09218} (\bibinfo{year}{2023}).
\newblock
\urldef\tempurl%
\url{https://doi.org/10.48550/ARXIV.2307.09218}
\showDOI{\tempurl}


\bibitem[Wang et~al\mbox{.}(2022)]%
        {SparCL}
\bibfield{author}{\bibinfo{person}{Zifeng Wang}, \bibinfo{person}{Zheng Zhan}, \bibinfo{person}{Yifan Gong}, \bibinfo{person}{Geng Yuan}, \bibinfo{person}{Wei Niu}, \bibinfo{person}{Tong Jian}, \bibinfo{person}{Bin Ren}, \bibinfo{person}{Stratis Ioannidis}, \bibinfo{person}{Yanzhi Wang}, {and} \bibinfo{person}{Jennifer Dy}.} \bibinfo{year}{2022}\natexlab{}.
\newblock \showarticletitle{SparCL: Sparse Continual Learning on the Edge}. In \bibinfo{booktitle}{\emph{Advances in Neural Information Processing Systems}}, Vol.~\bibinfo{volume}{35}. \bibinfo{pages}{20366--20380}.
\newblock


\bibitem[Zenke et~al\mbox{.}(2017)]%
        {SI}
\bibfield{author}{\bibinfo{person}{Friedemann Zenke}, \bibinfo{person}{Ben Poole}, {and} \bibinfo{person}{Surya Ganguli}.} \bibinfo{year}{2017}\natexlab{}.
\newblock \showarticletitle{Continual Learning Through Synaptic Intelligence}. In \bibinfo{booktitle}{\emph{Proceedings of the 34th International Conference on Machine Learning}} \emph{(\bibinfo{series}{Proceedings of Machine Learning Research}, Vol.~\bibinfo{volume}{70})}. \bibinfo{publisher}{PMLR}, \bibinfo{pages}{3987--3995}.
\newblock


\end{thebibliography}

\end{document}